\documentclass[sigconf]{acmart}
\AtBeginDocument{%
  \providecommand\BibTeX{{%
    \normalfont B\kern-0.5em{\scshape i\kern-0.25em b}\kern-0.8em\TeX}}}

\usepackage[ruled,vlined,linesnumbered]{algorithm2e}

\begin{document}

\title{ACGAN-GNNExplainer: Auxiliary Conditional Generative Explainer for Graph Neural Networks}

\author{Yiqiao Li}
\affiliation{%
  \institution{Data Science Institute \\
  University of Technology Sydney}
  \city{Sydney}
  \country{Australia}
  \postcode{2008}
}
\email{Yiqiao.Li-1@student.uts.edu.au}

\author{Jianlong Zhou}
\affiliation{%
  \institution{Data Science Institute \\
  University of Technology Sydney}
  \streetaddress{Ultimo}
  \city{Sydney}
  \country{Australia}}
\email{Jianlong.Zhou@uts.edu.au}

\author{Yifei Dong}
\affiliation{%
  \institution{Data Science Institute \\
  University of Technology Sydney}
  \city{Sydney}
  \country{Australia}
}
\email{Yifei.Dong@uts.edu.au}

\author{Niusha Shafiabady}
\affiliation{%
 \institution{Faculty of Science and Technology\\
 Charles Darwin University}
 \streetaddress{815 George Street}
 \city{Sydney}
 \country{Australia}}
 \email{Niusha.Shafiabady@cdu.edu.au}

\author{ Fang Chen}
\affiliation{%
  \institution{Data Science Institute\\
  University of Technology Sydney}
  \city{Sydney}
  \country{Australia}}
  \email{Fang.Chen@uts.edu.au}
  
\renewcommand{\shortauthors}{Yiqiao Li, Jianlong Zhou, Yifei Dong, Niusha Shafiabady, \& Fang Chen}




\begin{abstract}
Graph neural networks (GNNs) have proven their efficacy in a variety of real-world applications, but their underlying mechanisms remain a mystery. To address this challenge and enable reliable decision-making, many GNN explainers have been proposed in recent years. However, these methods often encounter limitations,
including their dependence on specific instances, lack of generalizability to unseen graphs, producing potentially invalid explanations, and yielding inadequate fidelity. To overcome these limitations, we, in this paper, introduce the Auxiliary Classifier Generative Adversarial Network (ACGAN) into the field of GNN explanation and propose a new GNN explainer dubbed~\emph{ACGAN-GNNExplainer}. Our approach leverages a generator to produce explanations for the original input graphs while incorporating a discriminator to oversee the generation process, ensuring explanation fidelity and improving accuracy. Experimental evaluations conducted on both synthetic and real-world graph datasets demonstrate the superiority of our proposed method compared to other existing GNN explainers.
\end{abstract}








\maketitle

\section{Introduction}\label{sec:intro}
Graph neural networks (GNNs) have emerged as a powerful tool for modelling graph-structured data and a natural choice for a variety of real-world applications such as crime prediction~\cite{wang2022hagen}, traffic forecasting~\cite{jiang2022graph}, and medical diagnosis~\cite{chen2021causal}, due to their ability to capture complex relationships between nodes and extract meaningful features from graphs. Notwithstanding its widespread adoption, its internal working mechanism remains a mystery, presenting potential challenges to its credibility and hindering its broader adoption in critical domains where explainability and transparency are essential. 

GNN explainers such as GNNExplainer~\cite{abs-1903-03894}, XGNN~\cite{yuan_xgnn:_2020}, and PGExplainer~\cite{LuoCXYZC020} have gained increasing attention in the field of explainable artificial intelligence (XAI), which attempts to identify the most important graph structures and/or features that contribute to GNNs' predictions. These methods have provided some elucidation of GNNs; however, substantial work is still required in the following aspects: 1)~\emph{Explanation scale} (local or global explanation): whether the explanation is linked to a specific instance or whether it has captured the archetypal patterns shared by the same group; 2)~\emph{Generalizability}: whether an explainer can be generalized to unseen graphs without retraining; 3)~\emph{Fidelity}: whether the explanations are real important subgraphs; 4)~\emph{Versatility}: whether an explainer is able to generate accurate explanations for different tasks such as node classification and graph classification. The pioneering method GNNExplainer~\cite{abs-1903-03894}, for instance, is limited to local explanation and lacks generalizability. Later, XGNN~\cite{yuan_xgnn:_2020} addressed this limitation but still lacks generalizability. Recent Gem~\cite{lin2021generative} has overcome the limitations of its predecessors, but the nature of its generation process makes its precision in explaining various tasks unstable.

In order to address the aforementioned challenges, we, in this paper, propose a new GNN explanation method dubbed ACGAN-GNNExplainer, which uses the auxiliary classifier Generative Adversarial Network (ACGAN)~\cite{ACGAN_OdenaOS17} as its backbone to generate explanations for graph neural networks. In particular, it consists of a generator and a discriminator. The generator learns to produce explanations based on these two pieces of information---the original graph $\mathbf{G}$ that requires an explanation and its corresponding label $f(\mathbf{G})$, which is determined by the target GNN model $f$. In addition, a discriminator is adopted to distinguish whether the generated explanations are "real" or "fake" and to designate a prediction label to each explanation. In this way, the discriminator could provide "feedback" to the generator and further monitor the entire generation process. Through iterative iterations of this interplay learning process between the generator and the discriminator, the generator ultimately is able to produce explanations akin to those deemed "real"; consequently, the quality of the final explanation is enhanced, and the overall explanation accuracy is significantly increased. Although ACGAN has been widely used in various domains (e.g., image processing~\cite{acgan_RoySSD18}, data augmentation~\cite{ZhouZT21}, medical image analysis~\cite{covidgan_abs}, etc.), to the best of our knowledge, this is the first time that ACGAN has been used to explain GNN models. Our method~\emph{ACGAN-GNNExplainer} has the following merits: 1) it learns the underlying pattern of graphs, thus naturally providing explanations on a goal scale; 2) after learning the underlying pattern, it can produce explanations for unseen graphs without retraining; 3) it is more likely to generate valid important subgraphs with the consistent monitoring of the discriminator; 4) it is capable of performing well under different tasks, including node classification and graph classification.

Our main contributions to this paper could be summarized as the following points:
\begin{itemize}
    \item We present a novel explainer, dubbed~\emph{ACGAN-GNNExplainer}, for GNN models, which employs a generator to generate explanations and a discriminator to consistently monitor the generation process;
    \item We empirically evaluate and demonstrate the superiority of our method ~\emph{ACGAN-GNNExplainer} over other existing methods on various graph datasets, including synthetic and real-world graph datasets, and tasks, including node classification and graph classification.
\end{itemize}

\section{Related Work}
\subsection{Generative Adversarial Networks}
Generative Adversarial Networks (GANs)~\cite{GAN_2014} are composed of two neural networks: a generator and a discriminator, trained in a game-like manner. The generator takes random noise as input and generates samples intended to resemble the training data distribution. On the contrary, the discriminator takes both real and generated samples as input and distinguishes between them. The generator tries to fool the discriminator by generating realistic samples, while the discriminator learns to accurately distinguish between real and fake samples. GANs have demonstrated successful applications across a wide range of tasks, including image generation, style transfer, text-to-image synthesis, and video generation.

Furthermore, the increasing utilization of GANs has led to the proposal of various variations, reflecting ongoing innovation and refinement within the field. These novel approaches introduce new architectural designs, optimization techniques, and training strategies to improve the stability, convergence, and overall quality of GAN models. 

Specifically, one strategy for expanding GANs involves incorporating side information. For instance, CGAN~\cite{CGAN_MirzaO14} proposes providing both the generator and discriminator with class labels to produce class conditional samples. Researchers in~\cite{PixelCNN_OordKEKVG16} demonstrate that class conditional synthesis significantly improves the quality of generated samples. Another avenue for expanding GANs involves tasking the discriminator with reconstructing side information. This is achieved by modifying the discriminator to include an auxiliary decoder network that outputs the class label of the training data or a subset of the latent variables used for sample generation. For example, Chen et al.~\cite{infoGAN_ChenCDHSSA16} propose InfoGAN, a GAN-based model that maximizes the mutual information between a subset of latent variables and the observations. It is known that incorporating additional tasks can enhance performance on the original task. In the paper~\cite{NguyenDYBC16}, the auxiliary decoder leverage pre-trained discriminators, such as image classifiers, to further improve the quality of synthesized images.

Motivated by the aforementioned variations, Odena et al.~\cite{ACGAN_OdenaOS17} introduce the Auxiliary Classifier Generative Adversarial Network (ACGAN), a model combining both strategies to leverage side information. Specifically, the proposed model is class-conditional, incorporating an auxiliary decoder tasked with reconstructing class labels.  ACGAN is an extension of CGANs. ACGANs are designed not only to generate samples that are similar to the training data and conditioned on the input information, but also to classify the generated samples into different categories. In ACGANs, both the generator and the discriminator are conditioned on auxiliary information, such as class labels. The generator takes random noise as input and generates samples conditioned on the input information and a set of labels, while the discriminator not only distinguishes between real and fake samples but also classifies them into different categories based on the input information.

ACGANs provide a way to generate diverse samples that are conditioned on the input information and classified into different categories, making them useful tools in many applications, such as image processing, data augmentation, and data balancing. In particular, the authors~\cite{acgan_RoySSD18} propose a semi-supervised image classifier based on ACGAN. Waheed et al.~\cite{covidgan_abs} apply ACGAN in medical image analysis. Furthermore, in~\cite{ZhouZT21}, authors augment the data by applying ACGAN in the electrocardiogram classification system. Ding et al.~\cite{fgcs_DingCDFC22} propose a tabular data sampling method that integrates the Knearest neighbour method and tabular ACGAN to balance normal and attack samples.
 \begin{figure*}[!htbp]
    \centering
\includegraphics[width=0.9\textwidth]{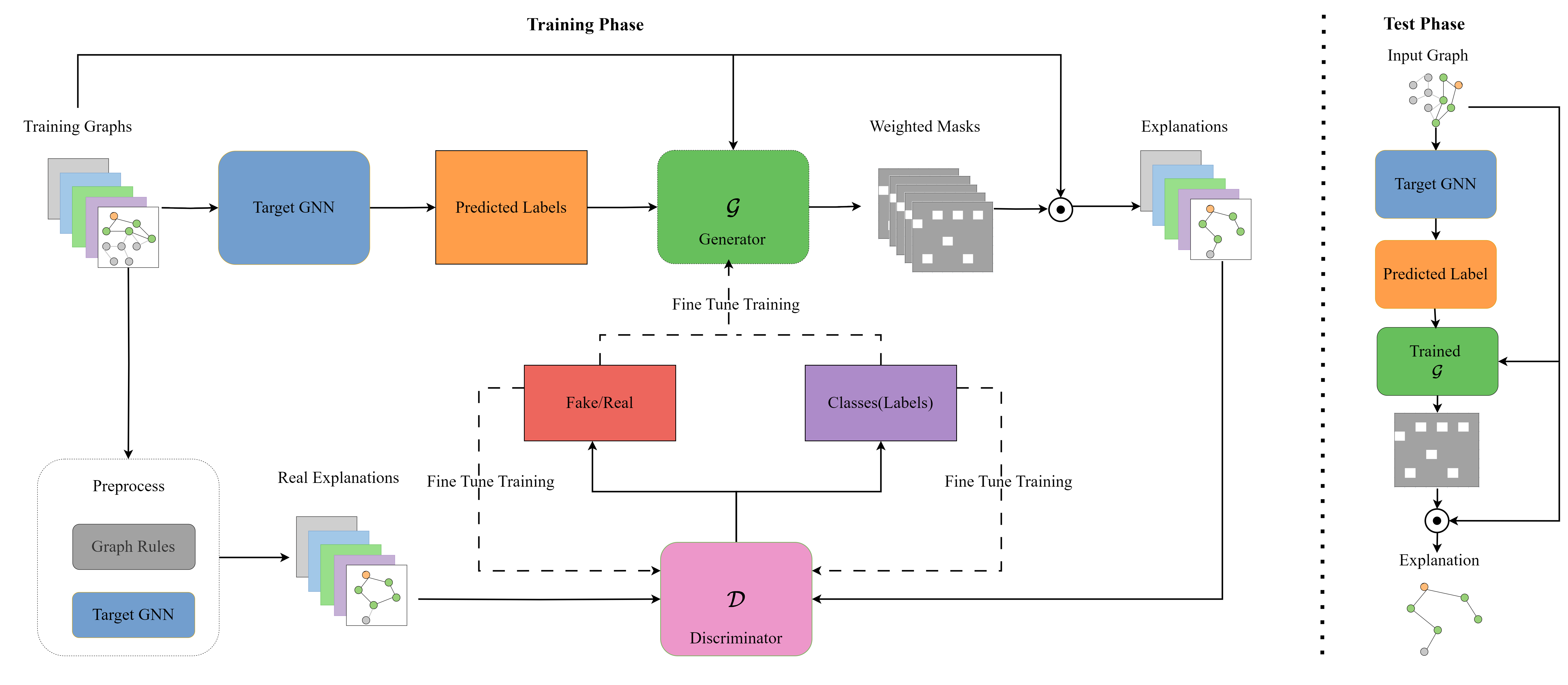}
    \caption{The framework of ACGAN-GNNExplainer. The $\odot$ means element-wise multiplication. This figure includes two phases: the training phase and the test phase. During the Training Phase, the objective is to train the generator and discriminator of the ACGAN-GNNExplainer model. After successful training, the Test Phase then utilizes the trained generator to generate explanations for the testing data.}
    \label{fig:framework}
\end{figure*}

\subsection{Graph Neural Network Explainers}
Explaining the decision-making process of graph neural networks (GNNs) is a challenging and important research topic, as it could greatly benefit users by improving safety and promoting trust in these models. To achieve this goal, several popular approaches have emerged in recent years that aim to explain GNN models by leveraging the unique properties of graph features and structures. In this regard, we briefly review several representative GNN explainers below.

\emph{GNNExplainer}~\cite{abs-1903-03894} is a seminal method in the field of explaining GNN models. By identifying the most relevant features and subgraphs that are essential to the predictions made by the GNN model, GNNExplainer is able to provide local-scale explanations for GNN models.~\emph{PGExplainer}~\cite{LuoCXYZC020} generates explanations for GNN models by using a probabilistic graph. Compared to GNNExplainer, it provides model-level explanations for each instance and has strong generalizability. Recent Gem~\cite{lin2021generative} offers local and global explanations and operates in an inductive configuration, allowing it to explain GNN models without retraining. Lin et al.~\cite{LinL0022_OrphicX} later propose OrphicX, which uses latent causal factors to generate causal explanations for GNN models. However, Gem~\cite{lin2021generative} and OrphicX~\cite{LinL0022_OrphicX} face difficulties in achieving consistently accurate explanations on real-world datasets. Therefore, in this paper, we attempt to develop a GNN explainer that is capable of generating explanations with high fidelity and precision for both synthetic and real-world datasets.

Furthermore, reinforcement learning is another prevalent technique employed for explicating GNN models. For example, Yuan et al.~\cite{XGNN_YuanTHJ20} propose XGNN, a model-level explainer that trains a graph generator to generate graph patterns to maximize a specific prediction of the model. Wang et al.~\cite {RC_Explainer_Wang} introduce RC-Explainer, which generates causal explanations for GNNs by combining the causal screening process with a Markov Decision Process in reinforcement learning. Furthermore, Shan et al.~\cite{RGExplainer_ShanSZLL21} propose RG-Explainer, a reinforcement learning enhanced explainer that can be applied in the inductive setting, demonstrating its better generalization ability.

In addition to the works we have mentioned above, there is another line of work that is working on generating counterfactual explanations. For example, the CF-GNNExplainer~\cite{DBLP_LucicHTRS22} generates counterfactual explanations for the majority of instances of GNN explanations. Furthermore, Bajaj et al.~\cite{BajajCXPWLZ21} propose an RCExplainer that generates robust counterfactual explanations, and Wang et al.~\cite{multi_grained_WangWZHC21} propose ReFine, which pursues multi-grained explainability by pre-training and fine tuning.  

\section{Method}
\subsection{Problem Formulation}
The notions of "interpretation" and "explanation" are crucial in unravelling the underlying working mechanisms of GNNs. Interpretation entails understanding the decision-making process of the model, prioritizing transparency, and the ability to trace the trajectory of decisions. In contrast, an explanation furnishes a rationale or justification for the predictions of GNNs, striving to present a coherent and succinct reasoning for the outcomes.

In this paper, we attempt to identify the subgraphs that significantly impact the predictions of  GNNs. We represent a graph as $\mathbf{G}=(\mathbf{V},\mathbf{A},\mathbf{X})$, where $\mathbf{V}$ is the set of nodes, $\mathbf{A} \in \left\{0,1\right\}$ denotes the adjacency matrix that $\mathbf{A}_{ij}=1$ if there is an edge between node $i$ and node $j$, otherwise $\mathbf{A}_{ij}=0$, and $\mathbf{X}$ indicates the feature matrix of the graph $\mathbf{G}$. We also have a GNN model $f$ and $\hat{l}$ denotes its prediction, $f(\mathbf{G}) \to \hat{l}$. We further define $E(f(\mathbf{G}), \mathbf{G})) \to \mathbf{G}^s$ as the explanation of a GNN explainer. Ideally, when feeding the explanation into the GNN model $f$, it would produce the exact same prediction $\hat{l}$, which means that $f(\mathbf{G})$ equals $f(E(f(\mathbf{G}), \mathbf{G}))$. We also expect that the explanation $E(f(\mathbf{G}), \mathbf{G})) \to \mathbf{G}^s$ should be a subgraph of the original input graph $\mathbf{G}$, which means that $\mathbf{G}^s \in \mathbf{G}$, so that the explained graph is a valid subgraph.

\subsection{Obtaining Causal Real Explanations}\label{subsec:casualty}
Our objective in this paper is to elucidate the reasoning behind the predictions made by the target GNN model $f$. To achieve this, we regard the target GNN model $f$ as a black box and refrain from investigating its internal mechanisms. Instead, we attempt to identify the subgraphs that significantly affect the predictions of the target GNN model $f$. In particular, we employ a generative model to autonomously generate these subgraphs/explanations. In order for the generative model to generate faithful explanations, it must first be trained under the supervision of "real" explanations (ground truth).  However, these ground truths are typically unavailable in real-world applications. In this paper, we employ Granger causality~\cite{granger2001investigating}, which is commonly used to test whether a specific variable has a causal effect on another variable, to circumvent this difficulty.

Specifically, in our experiments, we mask an edge and then observe its effect on the prediction of the target GNN model $f$. We then calculate the difference between the prediction probability of the original graph and the masked graph and set this difference as an edge weight to indicate its effect on the prediction of the target GNN model $f$. After that, we sort all edges of the graph according to the weight values we have obtained and save the resulting weighted graph. Therefore, edges with the highest weights correspond to actual explanations (important subgraphs). However, it should also be noted that using Granger causality~\cite{granger2001investigating} directly to explain a target GNN model $f$ is computationally intensive and has limited generalizability. Our method, on the other hand, could naturally overcome this challenge, as our parameterized explainer could capture the fundamental patterns shared by the same group and is adaptable and transferable across different graphs once the shared patterns have been comprehensively learned.

\subsection{ACGAN-GNNExplainer}\label{subsec_acgan_gnne}
Using the generating capacity of ACGAN, in this paper, we propose an ACGAN-based explanation method for GNN models, which is termed ACGAN-GNNExplainer. It consists of a generator ($\mathcal{G}$) and a discriminator ($\mathcal{D}$). The generator $\mathcal{G}$ is used to generate the explanations, while the discriminator $\mathcal{D}$ is used to monitor the generation process. The detailed framework of our method ACGAN-GNNExplainer is depicted in Figure~\ref{fig:framework}.

In contrast to the conventional strategy of training an ACGAN, in which random noise $\mathbf{z}$ is fed into the generator $\mathcal{G}$, our model feeds the generator $\mathcal{G}$ with the original graph $\mathbf{G}$, which is the graph we want to explain, and the label $L$, which is predicted by the target GNN model $f$. Employing this strategy, we ensure that the explanation produced by the generator $\mathcal{G}$, which plays a crucial role in determining the predictions of the GNN model $f$, corresponds to the original input graph $\mathbf{G}$. In addition, the generator $\mathcal{G}$ trained under this mechanism can be easily generalized to unseen graphs without significant retraining, thus saving computational costs. For the generator $\mathcal{G}$, we employ an encoder-decoder network where the encoder would project the original input graph $\mathbf{G}$ into a compact hidden representation, and the decoder would then reconstruct the explanation from the compact hidden representation. In our case, the reconstructed explanation is a mask matrix that indicates the significance of each edge.

Conceptually, the generator $\mathcal{G}$ is capable of generating any explanation (valid or invalid) if it is sufficiently complex, which contradicts the objective of~\emph{explaining} a GNN. Inspired by ACGAN, we adopt a discriminator $\mathcal{D}$ to monitor the generating process of $\mathcal{G}$. Specifically, our discriminator $\mathcal{D}$ is a graph classifier with five convolutional layers. It is fed with the real explanation and the explanation generated by our generator $\mathcal{G}$. It attempts to identify whether the explanation is "real" or "fake" and, at the same time classify the explanation, which serves as "feedback" to the generator $\mathcal{G}$ and further encourages the generator $\mathcal{G}$ to produce faithful explanations.

In addition, in order to train our generator $\mathcal{G}$ and discriminator $\mathcal{D}$, we need to obtain the "real" explanations first. To achieve this goal, we incorporate pre-processing in our framework (Fig.~\ref{fig:framework}), which uses the Granger causality~\cite{granger2001investigating} to acquire the "real" explanations. The details can be found in Section in~\ref{subsec:casualty}. Once the input graph $\mathbf{G}$, its corresponding real subgraphs (ground truth), and the labels have been acquired. We can train our ACGAN-GNNExplainer to produce a weighted mask that effectively highlights the edges and nodes in the original input graph $\mathbf{G}$ that significantly contribute to the decision-making process of the given GNN model $f$. Then, by multiplying the mask by the original adjacency matrix of the input graph, we obtain the corresponding explanations/important subgraphs. These explanations are particularly useful for comprehending the reasoning behind the complex GNN model. 

\subsection{Improved Loss Function}\label{subsec_loss}
The generator $\mathcal{G}$ generates explanations/subgraphs $\mathbf{G}^s \in \mathbf{G}$ based on two essential inputs: the original graph $\mathbf{G}$ and its associated label $l$, as expressed by $\mathbf{G}^s \leftarrow \mathcal{G}(\mathbf{G}, l)$. Concurrently, the discriminator $\mathcal{D}$ assesses the probabilities of origins ($S=\{$"$real$", "$fake$"$\}$) denoted as $P(S \mid \mathbf{G})$, and the probabilities of class classification (class labels $L=\{l_1,\cdots, l_n\}$) denoted as $P(L \mid \mathbf{G})$. Consequently, the loss function of the discriminator comprises two components: the log-likelihood of the correct source, $\mathcal{L}_S$, defined in Equation~\ref{eq:loss_D1}, and the log-likelihood of the correct class, $\mathcal{L}_L$, defined in Equation~\ref{eq:loss_D2}.

\begin{equation}\label{eq:loss_D1}
\begin{aligned}
 \mathcal{L}_S=
 & \mathbb{E}\left[\log P\left(S=\text{ real} \mid \mathbf{G} \right)\right]+ \\
& \mathbb{E}\left[\log P\left(S=\text { fake } \mid \mathbf{G^s}\right)\right] \\
\end{aligned}
\end{equation}

\begin{equation}\label{eq:loss_D2}
\begin{aligned}
 \mathcal{L}_L=
 & \mathbb{E}\left[\log P\left(L=l \mid \mathbf{G} \right)\right]+ \\
& \mathbb{E}\left[\log P\left(L=l \mid \mathbf{G^s} \right)\right]
\end{aligned}
\end{equation}

The discriminator $\mathcal{D}$ and the generator $\mathcal{G}$ play a minimax game, engaging in competitive interactions. The primary objective of the discriminator $\mathcal{D}$ is to maximize the probability of accurately classifying real and fake graphs ($\mathcal{L}_S$), as well as correctly predicting the class label ($\mathcal{L}_L$) for all graphs, resulting in a combined objective such as maximizing $(\mathcal{L}_{S} + \mathcal{L}_{L})$.

Conversely, the generator $\mathcal{G}$ aims to minimize the discriminator's capacity to identify real and fake graphs while simultaneously maximizing the discriminator's ability to classify them, as indicated by a combined objective like maximizing $(- \mathcal{L}_{S} + \mathcal{L}_{L})$. Thus, based on the Equation~\ref{eq:loss_D1} and Equation~\ref{eq:loss_D2}, the objective function of the $\mathcal{D}$ and $\mathcal{G}$ are formulated in Equation~\ref{eq:ACGAN_D_loss} and Equation~\ref{eq:ACGAN_G_loss}, respectively.

\begin{equation}\label{eq:ACGAN_D_loss}
\begin{aligned}
  \mathcal{L}_{(\mathcal{D})} =
& -\mathbb{E}_{\mathbf{G}^{gt} \sim P{(\mathbf{G}^{gt})}} \log \mathcal{D}(\mathbf{G}^{gt}) \\
& -\mathbb{E}_{\mathbf{G} \sim P{(\mathbf{G})}} \log [1-\mathcal{D}(\mathcal{G}(\mathbf{G}, l))] \\
 & - \mathbb{E}_{\mathbf{G}^{gt} \sim P{(\mathbf{G}^{gt})}} P(L \mid \mathbf{G}^{gt}) \\
 & -\mathbb{E}_{\mathbf{G} \sim P{(\mathbf{G})}} \log (P(L \mid \mathcal{G}(\mathbf{G}, l))
\end{aligned}
\end{equation}

\begin{equation}\label{eq:ACGAN_G_loss}
\begin{aligned}
 \mathcal{L}_{(\mathcal{G})}=
 & -\mathbb{E}_{\mathbf{G} \sim P{(\mathbf{G})}} \log \mathcal{D}(\mathcal{G}(\mathbf{G}, l)) \\
 & -\mathbb{E}_{\mathbf{G} \sim P{(\mathbf{G})}} \log P(L \mid \mathcal{G}(\mathbf{G}, l))
\end{aligned}
\end{equation}
Here, $\mathbf{G}$ represents the original graph that requires explanation, while $\mathbf{G}^{gt}$ signifies its corresponding actual explanation  (e.g., the real important subgraph). 

Using the objective functions as detailed in Equation~\ref{eq:ACGAN_D_loss} and Equation~\ref{eq:ACGAN_G_loss} for training the discriminator $\mathcal{D}$ and generator $\mathcal{G}$, it is observed that fidelity of the generated explanation is not satisfactory. This could be attributed to the fact that $\mathcal{L}_{(\mathcal{G})}$ as in Equation~\ref{eq:ACGAN_G_loss} does not explicitly incorporate the fidelity information from a target GNN model $f$. To overcome this limitation and enhance both the fidelity and accuracy of the explanation, we intentionally integrate the fidelity of the explanation into our objective function. Finally, we derive an enhanced generator ($\mathcal{G}$) loss function, as defined in Equation~\ref{eq:improved_ACGANloss}.

\begin{equation}\label{eq:improved_ACGANloss}
\begin{aligned}
\mathcal{L}_{(\mathcal{G})}=
& -\mathbb{E}_{\mathbf{G} \sim P{(\mathbf{G})}} \log \mathcal{D}(\mathcal{G}(\mathbf{G}, l)) \\
& -\mathbb{E}_{\mathbf{G} \sim p{(\mathbf{G})}} \log P(L \mid \mathcal{G}(\mathbf{G}, l)) \\
& +\lambda \mathcal{L}_{Fid}
\end{aligned}
\end{equation}

\begin{equation}\label{eq:fidlos}
\begin{aligned}
\mathcal{L}_{Fid} =  \frac{1}{N} \sum_{i=1}^N ||f(\mathbf{G})-f(\mathcal{G}(\mathbf{G}))||^2
\end{aligned}
\end{equation}
In this context, $\mathcal{L}_{Fid}$ represents the loss function component associated with fidelity. $f$ symbolizes a pre-trained target GNN model, $N$ signifies the count of node set of $\mathbf{G}$, and $\mathbf{G}$ represents the original graph intended for explanation. Correspondingly, $\mathbf{G}^{gt}$ stands for the explanation ground truth associated with it (e.g., the real important subgraph). Within this framework, $\lambda$ is a trade-off hyperparameter responsible for adjusting the relative significance of the ACGAN model and the explanation accuracy obtained from the pre-trained target GNN $f$. Setting $\lambda$ to zero results in Equation~\ref{eq:improved_ACGANloss} being precisely equivalent to Equation~\ref{eq:ACGAN_G_loss}. Notably, for our experiments, we selected $\lambda=2.0$ for synthetic graph datasets, $\lambda=4.0$ for the Mutagenicity dataset, and $\lambda=4.5$ for the NCI1 dataset.

\subsection{Pseudocode of ACGAN-GNNExplainer}
In Sections~\ref{subsec_acgan_gnne} and~\ref{subsec_loss}, we have described the framework and loss functions of our method in detail. To further elucidate our method, we provide its pseudocode in Algorithm~\ref{alg:ACGAN-GNNExplainer}. 

\begin{algorithm}[h]
\caption{Training a ACGAN-GNNExplainer}\label{alg:ACGAN-GNNExplainer}
\SetAlgoNoLine
\DontPrintSemicolon
\KwIn{Graph data $G=\{g_1,\cdots,g_n\}$ with labels $L=\{l_1,\cdots,l_n\}$, real explanations for graph data $G^{gt}=\{g_1^{gt},\cdots,g_n^{gt}\}$ (obtained in preprocessing phase), a pre-trained GNN model $f$}

\KwOut{A well-trained Generator $\mathcal{G}$, a well-trained discriminator $\mathcal{D}$}

Initialize the Generator $\mathcal{G}$ and the Discriminator $\mathcal{D}$ with random weights\;
\SetAlgoLined

\For{$epoch$ in $epochs$}{

    Sample a minibatch of $m$ real data samples $\{g^{(1)},\cdots,g^{(m)}\}$ and real labels $\{l^{(1)},\cdots,l^{(m)}\}$
    
    Generate fake data samples $\{g^{s(1)},\cdots,g^{s(m)}\} \leftarrow \mathcal{G}(g^{(1)},\cdots,g^{(m)})$\ and obtain their labels $\{l^{s(1)},\cdots,l^{s(m)}\}$
    
    Update $\mathcal{D}$ with the gradient:
    $$\nabla_{\theta_d}\frac{1}{m}\sum_{i=1}^m\left[\mathcal{L}_{(\mathcal{D})}\right]$$
 
    Update $\mathcal{G}$ with the gradient:
    $$\nabla_{\theta_g}\frac{1}{m}\sum_{i=1}^m\left[\mathcal{L}_{(\mathcal{G})}\right]$$
}
\end{algorithm}

\section{Experiments} 
In this section, we undertake a comprehensive evaluation of the performance of our proposed method, ACGAN-GNNExplainer. We first introduce the datasets we used in our experiments, as well as the implementation details in Section~\ref{subsec:imdetails}. After that, we show the quantitative evaluation of our method in comparison with other representative GNN explainers on synthetic datasets (see~Section~\ref{subsec:ressyn}) and real-world datasets (see Section~\ref{subsec:reseal}). Finally, we also provide a qualitative analysis and visualize several explanation samples generated by our method, as well as other representative GNN explainers in Section~\ref{subsec:visualization}.

\subsection{Implementation Details}\label{subsec:imdetails}
\paragraph{Datasets.} 
We focus on two widely used synthetic node classification datasets, including BA-Shapes and Tree-Cycles~\cite{abs-1903-03894}, and two real-world graph classification datasets, Mutagenicity~\cite{kazius2005derivation} and NCI1~\cite{WaleWK08}. Details of these datasets are shown in Table~\ref{tab:dataset}. 

\begin{table}
\centering
\caption{Details of synthetic and real-world datasets.}
\vspace{-10pt}
\setlength{\tabcolsep}{1.mm}{
\renewcommand\arraystretch{1.15}
\begin{tabular}{lcccc}
\toprule
& \multicolumn{2}{c}{Node Classification} & \multicolumn{2}{c}{Graph Classification}\\
\cmidrule(l){2-3}
\cmidrule(l){4-5}
& BA-Shapes & Tree-Cycles & Mutagenicity & NCI1\\
\midrule
\# of Graphs & 1 & 1  & 4,337 & 4110\\
\# of Edges & 4110 & 1950   & 266,894 & 132,753\\
\# of Nodes & 700  & 871   & 131,488 & 122,747\\
\# of Labels & 4 & 2  & 2 & 2\\    
\midrule
\end{tabular}
}
\label{tab:dataset}
\end{table}

\paragraph{Baseline Approaches.} Due to the growing prevalence of GNN in a variety of real-world applications, an increasing number of research studies seek to explain GNN, thereby enhancing its credibility and nurturing trust. Among them, we identify three representative GNN explainers as our competitors: \emph{GNNExplainer}~\cite{abs-1903-03894},~\emph{OrphicX}~\cite{LinL0022_OrphicX} and ~\emph{Gem}~\cite{lin2021generative}. For these competitors, we adopt their respective official implementations.

\begin{table*}
\caption{The fidelity and accuracy of explanations on BA-Shapes dataset: $Fid^+ (\uparrow)$, $Fid^- (\downarrow)$, $ACC_{exp} (\uparrow)$.}
\vspace{-10pt}
\centering
\setlength{\tabcolsep}{0.5mm}
{
\renewcommand\arraystretch{1.5}
\begin{tabular}{c|ccc|ccc|ccc|ccc|ccc}

\hline

\hline

 \multicolumn{1}{c|}{K}  & \multicolumn{3}{c|}{5} & \multicolumn{3}{c|}{6} & \multicolumn{3}{c|}{7} & \multicolumn{3}{c|}{8} & \multicolumn{3}{c}{9}\\

\cline{2-16}
(top edges) & $Fid^+$ & $Fid^-$ & $ACC_{exp}$ & $Fid^+$ & $Fid^-$ & $ACC_{exp}$ & $Fid^+$ & $Fid^-$ & $ACC_{exp}$ & $Fid^+$ & $Fid^-$ & $ACC_{exp}$ & $Fid^+$ & $Fid^-$ & $ACC_{exp}$ \\

\hline
GNNExplainer & 0.7059&	0.1471	&0.7941		&0.6765	&0.0588&	0.8824	&	0.7059	&0.0294	&0.9118	&	0.7353	&0.0000&	0.9412	&	0.7353	&0.0294	&0.9118 \\
Gem	& 0.5588&	\textbf{0.0000} &	\textbf{0.9412}	&	0.5588	& \textbf{-0.0294} &	\textbf{0.9706}	&	0.5882&	\textbf{-0.0294}	& \textbf{0.9706}	&	0.5882&	\textbf{-0.0294}	& \textbf{0.9706} &0.5882	& -0.0294 & 0.9706 \\
OrphicX	& \textbf{0.7941} &0.2059&	0.7353	&	\textbf{0.7941}	&0.2059& 0.7353	&	\textbf{0.7941}&	0.0882	&0.8529	& \textbf{0.7941}	&0.0588&	0.8824& \textbf{0.7941} &	0.0588	&0.8824 \\
Our Method	&0.6471	&0.1471	& 0.7941	&	0.5882	&0.0882	&0.8529	&	0.6176	& \textbf{-0.0294}	& \textbf{0.9706}	&	0.6471	& \textbf{-0.0294}	& \textbf{0.9706}	&	0.6471&	\textbf{-0.0588}	& \textbf{1.0000} \\

\hline

\hline
\end{tabular}
}
\label{tab:res_syn1}
\end{table*}

\begin{table*}
\caption{The fidelity and accuracy of explanations on Tree-Cycles dataset: $Fid^+ (\uparrow)$, $Fid^- (\downarrow)$, $ACC_{exp} (\uparrow)$.}
\vspace{-10pt}
\centering
\setlength{\tabcolsep}{0.5mm}
{
\renewcommand\arraystretch{1.5}
\begin{tabular}{c|ccc|ccc|ccc|ccc|ccc}

\hline

\hline

 \multicolumn{1}{c|}{K}  & \multicolumn{3}{c|}{6} & \multicolumn{3}{c|}{7} & \multicolumn{3}{c|}{8} & \multicolumn{3}{c|}{9} & \multicolumn{3}{c}{10}\\

\cline{2-16}

(top edges) & $Fid^+$ & $Fid^-$ & $ACC_{exp}$ & $Fid^+$ & $Fid^-$ & $ACC_{exp}$ & $Fid^+$ & $Fid^-$ & $ACC_{exp}$ & $Fid^+$ & $Fid^-$ & $ACC_{exp}$ & $Fid^+$ & $Fid^-$ & $ACC_{exp}$ \\

\hline

GNNExplainer	&0.9143	&0.8000	&0.1714	&	0.9429&	0.4571	&0.5143	&	\textbf{0.9714}	&0.1714	&0.8000	&	\textbf{0.9714}	&0.0571	&0.9143		& \textbf{0.9714}	&0.0571&	0.9143 \\
Gem&	\textbf{0.9714}	&0.2571	&0.7143	&	\textbf{0.9714}&	0.1429	&0.8286		& \textbf{0.9714}	&0.2571	&0.7143	&	\textbf{0.9714}	&0.1143	&0.8571	&	\textbf{0.9714}	&0.0857	&0.8857 \\
OrphicX	&0.9429	& \textbf{0.0000}	& \textbf{0.9714}	&	0.9429	& 0.0000&	0.9714&		0.9429	& \textbf{-0.0286}	& \textbf{1.0000}	&	0.9429	& \textbf{-0.0286}	& \textbf{1.0000}	&	0.9429	& \textbf{-0.0286} &	\textbf{1.0000} \\
Our Method& \textbf{0.9714} & \textbf{0.0000}	& \textbf{0.9714}	&	\textbf{0.9714}	& \textbf{-0.0286}	& \textbf{1.0000} &		\textbf{0.9714} &	0.0286	&0.9429&		\textbf{0.9714}	& 0.0571 &	0.9143	&	\textbf{0.9714} &	0.0000	& 0.9714 \\

\hline

\hline
\end{tabular}
}
\label{tab:res_syn4}
\end{table*}

\paragraph{Different Top Edges ($K$ or $R$).} After obtaining the weight (importance) of each edge for the input graph $\mathbf{G}$, it is also important to select the right number of edges to serve as the explanations as selecting too few edges may lead to an incomplete explanation/subgraph while selecting too many edges may introduce a lot of noisy information into our explanation. To overcome this uncertainty, we specifically define a top $K$ (for synthetic datasets) and a top $R$ (for real-world datasets) to indicate the number of edges we would like to select. We test different $K$ and $R$ to show the stability of our method. To be specific, we set $K=\{5,6,7,8,9\}$ for the BA-Shapes dataset, $K=\{6,7,8,9,10\}$ for the Tree-Cycles dataset, and $R=\{0.5, 0.6, 0.7, 0.8, 0.9\}$ for real-world datasets.

\paragraph{Data Split.} To maintain consistency and fairness in our experiments, we divide the data into three sets: 80\% for training, 10\% for validation, and 10\% for testing. Testing data remain untouched throughout the experiments. 

\paragraph{Evaluation Metrics.} A good GNN explainer should be able to generate concise explanations/subgraphs while maintaining high prediction accuracy when these explanations are fed into the target GNN. Therefore, it is desirable to evaluate the method with different metrics~\cite{yiqiao_survey}. In our experiments, we use the accuracy and fidelity of the explanation as our performance metrics. 

In particular, we generate explanations for the test set using \emph{GNNExplainer}~\cite{abs-1903-03894},~\emph{Gem}~\cite{lin2021generative},~\emph{OrphicX}~\cite{LinL0022_OrphicX}, and ACGAN-GNNExplainer (our method), respectively. We then feed these generated explanations to the pre-trained target GNN $f$ to compute the accuracy, which can be formally defined as Equation~\ref{eq:acc_exp}:
\begin{equation}\label{eq:acc_exp}
ACC_{exp}=\frac{|f(\mathbf{G})=f(\mathbf{G}^s)|}{|T|}
\end{equation}
where $\mathbf{G}$ signifies the initial graph necessitating explanation, and $\mathbf{G}^s$ denotes its associated explanation (e.g. the significant subgraph); $|f(\mathbf{G})=f(\mathbf{G}^s)|$ represents the count of accurately classified instances in which the predictions of $f$ on $\mathbf{G}$ and $\mathbf{G}^s$ are exactly the same, and $|T|$ is the total number of instances.

In addition, fidelity is a measure of how faithfully the explanations capture the important subgraphs of the input original graph. In our experiments, we employ the $Fidelity^+$ and $Fidelity^-$~\cite{yuan2021explainability} to evaluate the fidelity of the explanations.

$Fidelity^+$ quantifies the variation in the predicted accuracy between the original predictions and the new predictions generated by excluding the important input features. On the contrary, $Fidelity^-$ denotes the changes in prediction accuracy when significant input features are retained while non-essential structures are removed. Evaluation of both $Fidelity^+$ and $Fidelity^-$ provides a comprehensive insight into the precision of the explanations to capture the behaviour of the model and the importance of different input features. $Fidelity^+$ and $Fidelity^-$ are mathematically described in Equation~\ref{eq:fidelity+} and Equation~\ref{eq:fidelity-}, respectively.

\begin{equation}\label{eq:fidelity+}
Fid^+ = \frac{1}{N}\sum^{N}_{i=1} (f(\mathbf{G}_i)_{l_i}-f(\mathbf{G}_i^{1-s})_{l_i})
\end{equation}
\begin{equation}\label{eq:fidelity-}
Fid^- = \frac{1}{N}\sum^{N}_{i=1} (f(\mathbf{G}_i)_{l_i}-f(\mathbf{G}_i^{s})_{l_i})
\end{equation}
In these equations, $N$ denotes the total number of samples, and $l_i$ represents the class label for instance $i$. $f(\mathbf{G}_i)_{l_i}$ and $f(\mathbf{G}_i^{1-s})_{l_i}$ correspond to the prediction probabilities for class $l_i$ using the original graph $\mathbf{G}_i$ and the occluded graph $\mathbf{G}_i^{1-s}$, respectively. The occluded graph is derived by removing the significant features identified by the explainers from the original graph. A higher value of $Fidelity^+$ is preferable, indicating a more essential explanation. In contrast, $f(\mathbf{G}_i^{s})_{l_i}$ represents the prediction probability for class $l_i$ using the explanation graph $\mathbf{G}_i^{s}$, which encompasses the crucial structures identified by explainers. A lower $Fidelity^-$ value is desirable, signifying a more sufficient explanation.

\begin{table*}
\caption{The fidelity and accuracy of explanations on Mutagenicity dataset: $Fid^+ (\uparrow)$, $Fid^- (\downarrow)$, $ACC_{exp} (\uparrow)$.}
\vspace{-10pt}
\centering
\setlength{\tabcolsep}{0.5mm}
{
\renewcommand\arraystretch{1.5}
\begin{tabular}{c|ccc|ccc|ccc|ccc|ccc}

\hline

\hline

 \multicolumn{1}{c|}{R}  & \multicolumn{3}{c|}{0.5} & \multicolumn{3}{c|}{0.6} & \multicolumn{3}{c|}{0.7} & \multicolumn{3}{c|}{0.8} & \multicolumn{3}{c}{0.9}\\

\cline{2-16}

(edge ratio) & $Fid^+$ & $Fid^-$ & $ACC_{exp}$ & $Fid^+$ & $Fid^-$ & $ACC_{exp}$ & $Fid^+$ & $Fid^-$ & $ACC_{exp}$ & $Fid^+$ & $Fid^-$ & $ACC_{exp}$ & $Fid^+$ & $Fid^-$ & $ACC_{exp}$ \\

\hline
GNNExplainer & 0.3618 &	\textbf{0.2535} &	\textbf{0.6175}	&	0.3825&	0.2742&	0.5968	&	0.3963	&0.2396&	0.6313&		0.3641&	0.1774&	0.6935	&	0.3641	&0.0899	&0.7811 \\
Gem	&0.3018&	0.2972&	0.5737&		0.3295&	0.2696&	0.6014&		0.2857&	0.2120&	0.6590&		0.2581&	0.1475	&0.7235&		0.2120&	0.0806&	0.7903 \\
OrphicX	&0.2419&	0.4171	&0.4539	&	0.2949	&0.3111	&0.5599	&	0.2995&	0.2465&	0.6244		&0.3157&	0.1613	&0.7097		&0.2949	& \textbf{0.0599}	& \textbf{0.8111} \\
Our Method & \textbf{0.3963}	& \textbf{0.2535} &	\textbf{0.6175}	&	\textbf{0.3828} &	\textbf{0.2673}	&\textbf{0.6037}	&	\textbf{0.3986}	&\textbf{0.1636} &	\textbf{0.7074} &		\textbf{0.3602} &	\textbf{0.1037} &	\textbf{0.7673}	&	\textbf{0.3871}	&0.0806&	0.7903
  \\													
\hline

\hline
\end{tabular}
}
\label{tab:res_realworld_mutag}
\end{table*}

\begin{table*}
\caption{The fidelity and accuracy of explanations on NCI1 dataset: $Fid^+ (\uparrow)$, $Fid^- (\downarrow)$, $ACC_{exp} (\uparrow)$.}
\vspace{-10pt}
\centering
\setlength{\tabcolsep}{0.5mm}
{
\renewcommand\arraystretch{1.5}
\begin{tabular}{c|ccc|ccc|ccc|ccc|ccc}

\hline

\hline

 \multicolumn{1}{c|}{R}  & \multicolumn{3}{c|}{0.5} & \multicolumn{3}{c|}{0.6} & \multicolumn{3}{c|}{0.7} & \multicolumn{3}{c|}{0.8} & \multicolumn{3}{c}{0.9}\\

\cline{2-16}
(edge ratio) & $Fid^+$ & $Fid^-$ & $ACC_{exp}$ & $Fid^+$ & $Fid^-$ & $ACC_{exp}$ & $Fid^+$ & $Fid^-$ & $ACC_{exp}$ & $Fid^+$ & $Fid^-$ & $ACC_{exp}$ & $Fid^+$ & $Fid^-$ & $ACC_{exp}$ \\
\hline
GNNExplainer & 0.3358 & 0.2749 & 0.5961	& 0.3625 & 0.2603 & 0.6107 & 0.3844 & 0.1922 & 0.6788 & 0.3747 & 0.1095 & 0.7616 & 0.3236 & 0.0584 & 0.8127 \\
Gem	& 0.3796 & 0.3066 & 0.5645 & 0.4307& 0.2628 & 0.6083 & 0.4282 &	0.1873 & 0.6837 & 0.4404 & 0.1192 &	0.7518 & 0.3212& 0.0389 & 0.8321 \\
OrphicX	& 0.3114 & 0.3090 & 0.5620 & 0.3431	& 0.3236 & 0.5474 & 0.3382 & 0.2628 & 0.6083 & 0.3698 & 0.1630 & 0.7080 & 0.3139 & 0.0608 & 0.8102 \\
Our Method &	\textbf{0.4015}	& \textbf{0.2141} & \textbf{0.6569} & \textbf{0.4523} & \textbf{0.2214} &	\textbf{0.6496} & \textbf{0.4453} & \textbf{0.1849} &	\textbf{0.6861}	& \textbf{0.4672} & \textbf{0.0779} & \textbf{0.7932} & \textbf{0.3942} & \textbf{0.0254} & \textbf{0.8446} \\

\hline

\hline
\end{tabular}
}
\label{tab:res_realworld_nci1}
\end{table*}

In general, the accuracy of the explanation ($ACC_{exp}$) assesses the accuracy of the explanations, while $Fidelity^+$ and $Fidelity^-$ assess their necessity and sufficiency, respectively. A higher $Fidelity^+$ suggests a more essential explanation, while a lower $Fidelity^-$ implies a more sufficient one. Through comparison of accuracy and fidelity across different explainers, we can derive valuable insights into the performance and suitability of each approach.

\subsection{Experiments on Synthetic Datasets}\label{subsec:ressyn}

We first conduct experiments on two common synthetic datasets including BA-Shapes and Tree-Cycles~\cite{abs-1903-03894}, of which the details can be found in Section~\ref{subsec:imdetails}. We assess the fidelity and accuracy of the explanations generated by GNNExplainer, Gem, OrphicX, and our proposed ACGAN-GNNExplainer (our method). Table~\ref{tab:res_syn1} and Table~\ref{tab:res_syn4} present the fidelity and accuracy of explanations for BA-Shapes and Tree-Cycles datasets across different $K$, respectively. 

When examining the results for the BA-Shapes, as shown in Table~\ref{tab:res_syn1}, it is evident that no single model consistently surpasses the others across all metrics. However, as the value of $K$ increases, ACGAN-GNNExplainer progressively achieves competitive explanation accuracy $ACC_{exp}$ and better performance of $Fidelity^-$. On the contrary, OrphicX consistently exhibits higher $Fidelity^+$ values for various $K$, highlighting its proficiency in capturing essential subgraphs. However, its performance in terms of explanation accuracy $ACC_{exp}$ and $Fidelity^-$ lags behind, indicating that it struggles to provide comprehensive and precise explanations.  
\begin{figure*}[!htpb]
    \centering
\includegraphics[width=0.95\textwidth]{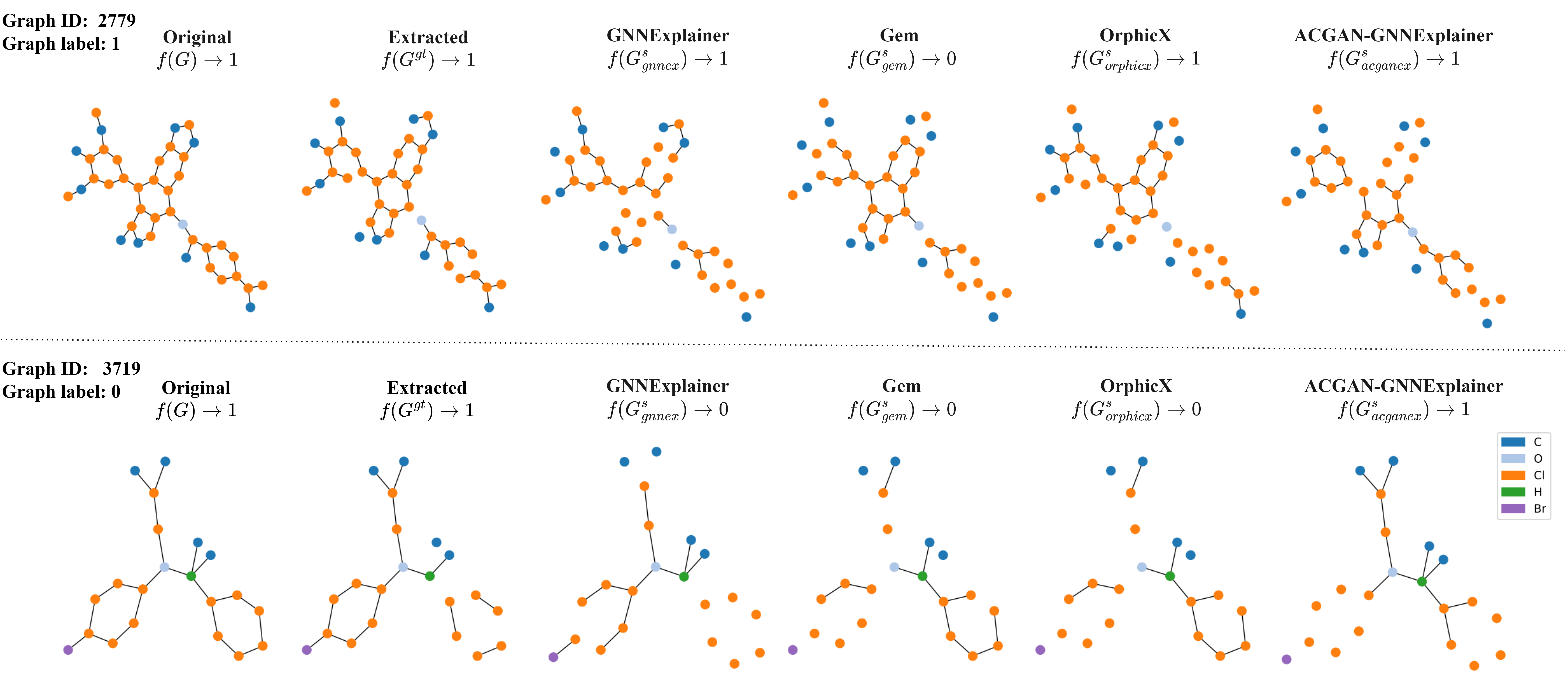}
    \caption{The explanation visualization on NCI1 when $R=0.5$. $f(\cdot) \to \{0,1\}$ means predictions made by the target GNN model $f$. The $1^{st}$ column contains the initial graph. The $2^{nd}$ column showcases the real explanation that we obtained during the preprocessing stage. The $3^{rd}$ to $5^{th}$ columns are the explanations produced by GNNExplainer, Gem, OrphicX and ACGAN-GNNExplainer, respectively. On analyzing the first row, we observe that GNNExplainer, OrphicX, and ACGAN-GNNExplainer successfully obtain the explanations that are successfully classified by the target GNN model $f$. However, upon examining the visualization of the explanation subgraph, it is obvious that the explanation produced by ACGAN-GNNExplainer exhibits the closest resemblance to the real explanations. Moving on to the second row, we find that ACGAN-GNNExplainer tends to select molecules other than the Cl circle as part of the explanation subgraph. In contrast, other competitors have a tendency to include the Cl molecule circle as part of the explanation subgraph.}
    \label{fig:res_nci1}
\end{figure*}
Upon analyzing the results presented in Table~\ref{tab:res_syn4}, it is evident that all methods demonstrate a commendable performance on the Tree-Cycles datasets with different $K$ values. However, no single method consistently outperforms the others in all evaluation metrics, which shows a trend similar to the results in the BA shapes (see Table~\ref{tab:res_syn1}). Notably, within the range of $K=\{6, 7\}$, ACGAN-GNNExplainer emerges as the superior choice among all the alternatives. It maintains the highest fidelity compared to the other methods on all $K$ values. Although outperformed by Orphicx in terms of $Fidelity^-$ and accuracy $ACC_{exp}$ when $K$ is in the range of $\{8, 9, 10\}$, ACGAN-GNNExplainer still shows competitive performance. 

In summary, all GNN explainers manifest robust performance in synthetic datasets, largely attributed to their intrinsic simplicity in contrast to real-world datasets. Notably, ACGAN-GNNExplainer consistently outperforms alternative methods in several scenarios. Moreover, even in situations where ACGAN-GNNExplainer does not outshine its counterparts, it maintains competitive levels of performance. To offer a comprehensive evaluation of ACGAN-GNNExplainer, we extend our exploration to real-world datasets in the forthcoming Section~\ref{subsec:reseal}, facilitating a thorough analysis.

\subsection{Experiments on Real-world Datasets}\label{subsec:reseal}
Here we further experiment with our method with two popular real-world datasets including Mutagenicity~\cite{kazius2005derivation} and NCI1~\cite{WaleWK08}. The experimental results for Mutagenicity and NCI1 are shown in Table~\ref{tab:res_realworld_mutag} and Table~\ref{tab:res_realworld_nci1}, respectively.

From Table~\ref{tab:res_realworld_mutag}, it can be seen that ACGAN-GNNExplainer demonstrates superior performance in both fidelity ($Fidelity^+$, $Fidelity^-$) and accuracy $ACC_{exp}$ in most settings where $R$ ranges from $0.5$ to $0.8$. While OrphicX marginally outperforms ACGAN-GNNExplainer in terms of explanation accuracy $ACC_{exp}$ when $R=0.9$, its fidelity lags behind. However, maintaining high fidelity without sacrificing accuracy is crucial when explaining GNNs in practice. From this perspective, our method shows an obvious advantage over others. Similarly, from Table~\ref{tab:res_realworld_nci1}, one can observe that ACGAN-GNNExplainer consistently outperforms other competitors in terms of fidelity and accuracy in different $R$.

Our method consistently yields higher $Fidelity^+$ scores, suggesting that our generated explanations have successfully covered the important subgraphs. On the other hand, our method achieved lower $Fidelity^-$ scores compared to other methods. This highlights the sufficiency of our explanations, as they effectively conveyed the necessary information for accurate predictions while mitigating inconsequential noise. Furthermore, in terms of accuracy, our method consistently yields higher explanation accuracy compared with other methods, underscoring its proficiency in effectively capturing the underlying rationale of the GNN model. In general, these results highlight the effectiveness of our proposed method in producing faithful explanations.

\subsection{Qualitative Analysis}\label{subsec:visualization}
Qualitative evaluation is another effective way to compare explanations generated by different explainers. Here, we present visualizations of the explanations on NCI1 with $R=0.5$ and visualize two examples of explanations---the target GNN model $f$ successfully classifies one example but fails to classify the other one. We try to investigate the factors that affect the predictions of the target GNN model $f$---resulting in a correct prediction or causing a wrong prediction. Specifically, when the target GNN model $f$ yields a correct prediction (e.g., the first-row visualization example in Figure~\ref{fig:res_nci1}), our objective is to provide an explanation that would highlight the key elements that lead to the correct prediction. Conversely, when the target GNN model $f$ produces an incorrect prediction (e.g., the second-row visualization example in Figure~\ref{fig:res_nci1}), we hope to offer an explanation that elucidates the factors contributing to the incorrect prediction.

Therefore, our goal is to ensure that the explanation generated by our proposed method aligns well with the prediction made by the target GNN model $f$. In particular, when the target GNN model $f$ accurately predicts the label for a given graph, we expect our explanation to yield the same prediction. As illustrated in the first row of Figure~\ref{fig:res_nci1}, we observe that GNNExplainer, Orphicx, and ACGAN-GNNExplainer provide correct explanations for the graph that the GNN correctly predicts. However, it is worth noting that the explanation subgraph generated by ACGAN-GNNExplainer exhibits the closest resemblance to the real explanation subgraph extracted in the preprocessing phase. Furthermore, when examining another graph for which the target GNN model $f$ makes an incorrect prediction, we find that only ACGAN-GNNExplainer is capable of producing a correct explanation. Notably, the ACGAN-GNNExplainer model demonstrates a tendency to select other molecules as part of the explanation subgraph rather than the Cl circles. In contrast, other methods we have compared tend to include the Cl molecule circle as part of the explanation subgraph.

Visually, our method demonstrates a higher degree of visual similarity to the actual explanation in comparison to other competing methods. This observation provides additional evidence supporting the efficacy of our method in producing faithful explanations.

\section{Conclusion}
Unboxing the intrinsic operational mechanisms of a GNN is of paramount importance in bolstering trust in model predictions, ensuring the reliability of real-world applications, and advancing the establishment of trustworthy GNNs. In pursuit of these objectives, many methods have emerged in recent years. Although they demonstrate commendable functionality in certain aspects, most of them struggle in obtaining good performance on real-world datasets.

To address this limitation, we, in this paper, propose an ACGAN-based explainer, dubbed ACGAN-GNNExplainer, for graph neural networks. This framework comprises a generator and a discriminator, where the generator is used to generate the corresponding explanations for the original input graphs and the discriminator is used to monitor the generation process and signal feedback to the generator to ensure the fidelity and reliability of the generated explanations. To assess the effectiveness of our proposed method, we conducted comprehensive experiments on synthetic and real-world graph datasets. We performed fidelity and accuracy comparisons with other representative GNN explainers. The experimental findings decisively establish the superior performance of our proposed ACGAN-GNNExplainer in terms of its ability to generate explanations with high fidelity and accuracy for GNN models, especially on real-world datasets.

\balance
\bibliographystyle{ACM-Reference-Format}
\bibliography{reference}

\end{document}